\documentstyle{elsart}
\newtheorem{theorem}{\sc Theorem}
\newtheorem{lemma}{\sc Lemma}
\newtheorem{prope}{\sc Property}
\newtheorem{coro}{\sc Corollary}
\newtheorem{nota}{\sc Notation}
\newtheorem{defin}{\sc Definition}

\newtheorem{ex}{\sc Example}

\newenvironment{proof}{\par \sc Proof.\rm}{\hspace*{\fill}$\Box$\vspace{1ex}}
\newenvironment{example}{\begin{ex}}{\hspace*{\fill}$\Diamond$\end{ex}}
\newenvironment{corollary}{\begin{coro}}{\end{coro}}
\newenvironment{property}{\begin{prope}}{\end{prope}}
\newenvironment{definition}{\begin{defin}}{\end{defin}}

\begin{document}
\date{}
\begin{frontmatter}
\title{A Discipline of Evolutionary Programming\thanksref{title}}
\author{Paul Vit\'anyi\thanksref{author}}
\address{CWI, Kruislaan 413, 1098 SJ Amsterdam,
The Netherlands. Email: paulv@cwi.nl; WWW: http://www.cwi.nl/$\sim$paulv/}

\thanks[title]{Preliminary version published in:
{\em Proc. 7th Int'nl Workshop on Algorithmic Learning Theory,
       Lecture Notes in Artificial Intelligence}, Vol. 1160,
  Springer-Verlag, Heidelberg,
1996, 67-82.} 
\thanks[author]{Partially
supported by the European Union
through NeuroCOLT ESPRIT Working Group Nr. 8556,
and by  NWO through NFI Project ALADDIN under Contract
number NF 62-376. 
Author's affilliations are CWI and the University of Amsterdam.}

\begin{abstract}
Genetic fitness optimization using small populations or small
population updates across generations
generally suffers from randomly diverging evolutions.
We propose a notion of
highly probable fitness optimization through 
feasible evolutionary computing runs 
on small size populations.
Based on rapidly mixing Markov chains, the approach
pertains to most types of evolutionary
 genetic algorithms, genetic programming and the like.
We establish that for systems having associated rapidly mixing
Markov chains and appropriate stationary distributions
the new method finds optimal programs (individuals) with probability
almost 1. To make the method useful would require
a structured design methodology where
the development of the program and the guarantee
of the rapidly mixing property go hand in hand.
We analyze a simple example to show that the method
is implementable. More significant examples require theoretical
advances, for example with respect to the Metropolis filter.
\end{abstract}
\end{frontmatter}

\section{Introduction}
Performance analysis of genetic computing using unbounded
or exponential population sizes or population updates across generations
\cite{VoLi91,RSW92,SrPa93,VoWr94,WrVo95,RRS95,Go95}
may not be directly
applicable to real practical problems where we always
have to deal with a bounded (small) population size \cite{Go89,Re93,Su93}.

Considering small population sizes it is at once obvious
that the size and constitution of the population or population
updates may have
a major impact on the evolutionary development of the population.
We aim to establish a fast feasible speed of convergence
to a distribution of populations from which we can obtain by Monte Carlo
sampling an optimal type individual with high probability.
The method we propose clearly can be used by a wide range
of genetic computing models which includes genetic algorithms
on strings and genetic programming on trees, and so forth.
Application of the method to problems is another matter;
we have examples solving trivial problems
but we don't have an example solving a difficult problem.
The main question for future research is to supply such an application.

The structure of the paper is as follows.
In Section~\ref{sect.model} we explain
the finite Markov chain model
for genetic processes. The states of the chain
correspond to finite populations. The transition
probability between two states is induced by the
selection, reproduction, and fitness rules, as
in \cite{NV92,Su93,JSG94}.

Since the evolution from generation to generation is a random
process, using finite populations different evolutions may diverge.
This is not the case when we consider evolutions
of probability density distributions. The idea is to view such
processes as corresponding with infinite populations that
are completely transformed in each generation. Or to view them
as an approximation to very large populations with very large updates
between generations or as an average of all possible evolutions from
a finite population. Such evolutions considered by
several authors as a convenient vehicle to analyze genetic processes
are completely deterministic. In Section~\ref{sect.evol} we show
that even if we view such a deterministic evolution as an ``average''
evolution, this average may behave very different from every particular
real evolution. The crucial point here is how far a particular evolution
(generally) strays from the average. We analyze the relation with the
population size and the population update size.

Under mild conditions that guarantee ergodicity the Markov chain converges
to a stationary distribution over the set of states
(the set of reachable populations).
From this stationary distribution we can
sample a set of populations. If the total stationary probability
concentrated on populations containing
an individual of best fitness is large enough
then this process finds such an individual with high probability.
For this approach to be workable we must have small enough
populations and the convergence to the stationary
distribution has to be fast. 
Convergence to stationarity is fast enough in ``rapidly mixing''
Markov chains. Such chains have recently been the basis
of spectacular randomized approximation algorithms,
combinatorial counting, statistical physics,
combinatorial optimization, and certain quadratic dynamic processes
related to genetics of infinite populations,
\cite{Si92,DS91,Fi91,RRS95}. 
Section~\ref{sect.rapmix} introduces them in
general genetic computing. 
The efficiency of our technique in applications depends crucially
on the rate of convergence of the Markov chain. Since the number
of states is typically very large, the chain should reach equilibrium
after each particular evolution has only explored a tiny fraction
of the state space. 

For the {\em theory} of genetic computing it
is important that we demonstrate a
formal method of genetic fitness optimization (applicable to
restricted classes of GA, GP, and related optimization problems) 
together with a rigorous analysis demonstrating
that this strategy is {\em guaranteed} to work with {\em high probability},
rather than intuitive heuristic or ad hoc arguments.
For the {\em application} of genetic computing we find that
because of the sampling from the stationary distribution the proposed process
uses a large number of short runs 
as opposed to one long run. 
\footnote{From a more applied perspective
several researchers observed earlier that it pays 
to restart on a new population when the evolution 
takes a unpromising direction, for example \cite{JS89,Go89}.
Also J. Koza and  L.J.  
Eshelman have algorithms that specifically restart
automatically (GP, CHC, respectively), as do many others.}

Just to show that the method is meaningful we demonstrate it
 on a toy problem in Section~\ref{sect.toy} that in fact is trivially
successful because of the abundance of optimal solutions.
Really significant examples are currently much harder---and already beyond
the scope of this exploration. Further along is
the development of a structured methodology
to set up the genetic system (selection, reproduction,
fitness) such that the resulting Markov chain is rapidly mixing,
and, moreover, such that the types with sufficiently high fitness
will be obtained with sufficiently
high probability from the (close to) final stationary
state distribution. What we have in mind is a design
methodology to develop a genetic system satisfying
these requirements from the specifications of the problem statement.

\section{The Model}\label{sect.model}
Assume that $r$ is an upper bound on the
number of different possible types of individuals,
say a set $\Omega =\{0, \ldots , r-1\}$. Such individuals can be strings, trees
or whatever---our discussion is so general that the precise
objects don't matter. Even if the set of types can grow (such as trees)
then practically speaking there will still be an upper bound $r$. 
The genetic system tries to solve an optimization
problem in the following sense. 
Each individual in $\Omega$ is graded in terms of
how well it solves the problem the genetic system is supposed to solve,
expressed as a function $f$ which maps $\Omega$ to some grading set $G$.
For example, $G$ can be the real interval $[0,1]$.
Let $f(u)$ be the {\em fitness} of type $u$. Then,
the normalized fitness
of individual $u$ is
\[ \hat f(u) = {{f(u)} \over {\sum_{v \in \Omega} f(v) }}. \]
To fix thoughts, we use fitness proportional selection where
selection of individuals from a population
is according to probability related to the product
of frequency of occurrence and fitness. That is, in
a population $P=(P(1), \ldots , P(r))$ of size $n$,
where type $u$ occurs with
frequency $P(u) \geq 0$ with $\sum_{u \in \Omega} P(u)=n$, 
we have probability $p(u)$ to select
individual $u$ (with replacement) for the cross-over
defined by
\[ p(u) = {{ f(u) P(u)} \over {\sum_{v \in \Omega} f(v)P(v) }}.\]
It is convenient to formulate the generation of one population from
another one as a Markov chain. Formally:
\begin{definition}
A sequence of random variables $(X_t)_{t=0}^{\infty}$ with
outcomes in a finite state space $T=\{0, \ldots , N-1\}$
is a {\em finite state time-homogeneous Markov chain}
if for every ordered pair $i,j$ of states the quantity
$q_{i,j} = \Pr (X_{t+1} = j | X_t = i)$ called the 
{\em transition probability} from state $i$ to state $j$,
is independent of $t$. If ${\cal M}$ is a Markov chain
then its associated {\em transition matrix} $Q$ is defined as
$Q:=(q_{i,j})_{i,j=0}^{N-1}$. The matrix $Q$ is
non-negative and {\em stochastic}, its row sums are all unity.
\end{definition}
Now let the Markov chain ${\cal M}$ have
states consisting of nonnegative integer $r$-vectors
of which the individual entries sum up to the population size exactly $n$
and let ${\cal P}$ denote the set of states of ${\cal M}$. The 
number of states $N := \#{\cal P}$ is given by \cite{NV92}
\begin{equation}\label{eq.mcstates}
N =  {{n+r-1} \choose {r-1} }.
\end{equation}
(This is the number of ways we can select $r-1$ elements from $n+r-1$
elements. If the elements constitute a linear list and the $r$ 
intervals marked by the 
selected elements---exlusive the selected elements---represents
the elements of the $r$ types
the result follows directly.)
The associated transition matrix $Q=(q_{i,j})$
is a $N \times N$ matrix where the entry $q_{i,j}$ is the probability
that the $k$th generation will be $P_j$ given that the $(k-1)$st
generation is $P_i$ ($P_i,P_j \in {\cal P}$).

A general closed form expression for transition
probabilities for simple GA's is derived in \cite{NV92}
and its asymptotics to steady state distributions as
population size increases is determined. 
In \cite{JSG94} it is observed that the mentioned closed form expression 
allows expression of `expected waiting
time until global optimum is encountered for the first time',
`expected waiting time for first optimum within some error tolerance
of global optimum', and `variance in such measures from run to run',
and so on, but no further analysis is provided. Instead,
initial experimental work is reported. Here we are
interested in quantitative estimates of such expressions.

\begin{example}\label{ex.small}
\rm
Consider a process where the generation
of a next population $P'$ from the current
population $P$ consists of sampling two individuals $u,v$
from $P$, removing these two individuals
from $P$ ($P'':= P- \{u,v\}$), producing two new offspring $w,z$ and
inserting them in the population
resulting in a population $P' := P'' \bigcup \{w,z\}$.

The transition probability $q_{P,P'}$ of
\[ P \rightarrow P'\]
where $P'$ results from sequentially executing
the program
``$P(u):=P(u)-1$; $P(v) :=P(v)-1$;
$P(w):=P(w)+1$; $P(z): =P(z)+1$; $P' := P$,''
replacing
the pair of individuals $u,v$ by $w,z$ with 
$\{u,v\} \bigcap \{w,z \} = \emptyset$, is given by
\begin{equation}\label{eq.trans}
q_{P,P'} := 2 p(u)p(v)b(u,v,w,z),
\end{equation}
where the {\em local transition probability} $b(u,v,w,z)$ is the probability of producing the pair $w,z$
from the selected pair $u,v$, incorporating both the mutation probability
and the cross-over probability.
The $r \times r \times r \times r$ matrix $B=(b(u,v,w,z))$ is called
the {\em local transition matrix}.
We can generally obtain such
transition probabilities between states
$ P \rightarrow P'$ of the Markov chain. 
\end{example}
 
\section{Evolutionary Trajectories}\label{sect.evol}
Small population size or sample size may cause 
evolutionary trajectories to drift apart. Without
loss of generality, this can be illustrated
in a simplified setting ignoring fitness selection.

\subsection{Transformation of Distributions}
Some approaches use the expedient to  simply ignore the actual populations
and deal with the probability density $p( \cdot)$ of types rather than with
the number of occurrences $P(\cdot)$ of types in a population $P$.
We show that the idea that the deterministic evolution is some sort
of ``average'' of all evolutions of underlying populations has
problems. 
Given a distribution density $p$ and a local
transition matrix $B=(b(u,v,w,z))$, let
the transformation $p'=g(p)$ be defined by
\begin{equation}\label{eq.Btransform}
 p' (z) := \sum_{u,v} \left( p(u)p(v) \sum_w b(u,v,w,z) \right) ,
\end{equation}
where $B$ is such that $p'$ is again a probability
distribution.
Consider a (not necessarily finite)  population of individuals, each individual
being of some type $u \in \{0, \ldots , r-1\}$.
Let $p(u)$ be the probability of selecting
an individual of type $u$. When a pair of individuals of types $u,v$
mate then they 
produce  a pair of individuals of types $w,z$ with probability
$b(u,v,w,z)$. Assuming that a mating of a pair must
result in a pair of offspring
means that $\sum_{w,z} b(u,v,w,z)=1$.
The resulting probability of $z$ is $p'(z)$ in Equation~\ref{eq.Btransform}.
Then,
\begin{eqnarray*}
&& \sum_{u,v} p(u)p(v) b(u,v,w,z)  =  p'(w)p'(z) \\
&& \sum_{w,z} p'(w)p'(z)  =  1 .
\end{eqnarray*}

Probability density evolution has particular nice properties
that can be demonstrated not to hold for population evolutions.
In particular, probability density evolution converges.
A distribution $\rho$ is called an {\em equilibrium distribution}
(with respect to transformation $g$) if $g(\rho)=\rho$.
In \cite{DP91,NV92} for simple GA with fitness selection,
and \cite{RSW92} for more general quadratic dynamical systems
but without fitness selection, the following convergence property is
derived.
\begin{theorem}\label{theorem.RSW}
The sequence $p^0, p^1, \ldots$
with $p^t = g^t(p^0)$ ($t \geq 0$)
converges to an equilibrium distribution 
$\lim_{t \rightarrow \infty} p^t = \rho$.
\end{theorem}
In certain infinite evolutionary models equivalent to the above transformation
of probability densities the
evolution develops deterministically according to
Equation~\ref{eq.Btransform}. But in practice things are different.
Namely, the single evolution of the probability density
may be very different from {\em every} evolution of a represented population.

If the populations
are small, or the population updates across successive
generations are small, then we are dealing with a random
process and chance selections can cause great
divergence of evolution of populations. In the practice
of evolutionary computing this is always the case. We would like
to quantify this. Primarily considering 
probability densities, neither \cite{NV92}
nor \cite{RSW92} explores in an explicit 
quantitative manner the divergence of trajectories
of individual runs based on population sizes.
They rather focus on the issue that as the population size grows,
the divergence of possible trajectories gets progressively 
smaller. In the limit, for infinite
populations, the generations in the run converge to the 
expected trajectory for smaller populations. 
Clearly, if all trajectories are in a small envelope around
the expected trajectory, then the expected trajectory is a good
predictor for what happens with an individual run. If moreover the expected
trajectory corresponds to the trajectory of Equation~\ref{eq.Btransform}, 
as in the system
analyzed in \cite{NV92}, then the analysis
of transformation $g$  tells us what to expect from our individual
bounded population evolution. 

However, the {\em expected trajectory}
can be {\em completely different} from {\em all individual trajectories};
and if the individual trajectories
of bounded populations diverge wildly,
then the expected trajectory may not predict anything about
what happens to an individual run.
Analysis like in \cite{NV92,RSW92} do not
deal with an individual run of a genetic
algorithm, but rather with the sequence
of expectations over all individual runs of the system.
Such an expectation may not say anything about
what actually happens.

To see this, consider a {\em dictatorial coin}
which gives a first outcome 0 or 1 with fair
odds. However, afterwards it always gives the same outcome.
So it either produces an all 0 run or an all 1 run
with equal probabilities. The expectation of obtaining
a 0 at the $t$th trial is $1/2$. However, in actual
fact at the $t$th ($t>1$) trial we have either probability 1
or probability 0 for
outcome 0. In terms of the above formalism,
initially, $p(0)=p(1)=1/2$.
To express the ``dictatorial coin'' in terms of evolutionary
processes and to analyze what happens we continue the above
Markov chain termonology.

For $s \in {\cal N}$, the $s$-{\em step transition matrix} is the power
$Q^s = (q_{i,j}^s)$ with $q_{i,j}^s= \Pr(X_{t+s}=j|X_t=i)$,
independent of $t$. Denote the distribution of $X_t$ by
the row vector $\pi^t = ( \pi_0^t , \ldots , \pi_{N-1}^t)$
with $\pi_i^t = \Pr(X_t=i)$. If $\pi^0$ denotes the initial
distribution then $\pi^t = \pi^0 Q^t$ for all $t \in {\cal N}$.
Often we have $\pi_i^0 =1$ for some $i$ (and 0 elsewhere) in
which case $i$ is called the {\em initial state}.

\begin{definition}
The chain is {\em ergodic} if there exists a distribution $\pi$
over ${\cal P}$ with strictly positive probabilities such that
\[ \lim_{s \rightarrow \infty} p_{i,j}^s = \pi_j , \]
for all $P_i,P_j \in {\cal P}$. In this case we have that
$\pi^t = \pi^0 Q^t \rightarrow \pi$ pointwise
as $t \rightarrow \infty$, and the limit is independent of $\pi^0$.
The {\em stationary distribution} $\pi$ is the unique vector
satisfying $\pi Q = \pi$, where $\sum_i \pi_i =1$; that is,
the unique normalized left eigenvector of $Q$ with eigenvalue 1.
Necessary and sufficient conditions for ergodicity are that
the chain should be {\em irreducible}, 
for each pair of states $P_i,P_j \in {\cal P}$
there is an $s \in {\cal N}$ such that $p_{i,j}^s > 0$ ($P_j$
can be reached from $P_i$ in a finite number of steps);
and {\em aperiodic}, the $\mbox{gcd}\{s:p_{i,j}^s>0\}=1$ for all
$P_i,P_j \in {\cal P}$.
\end{definition}
An ergodic Markov chain is ({\em time-}){\em reversible} iff either
(and hence both) of the following equivalent conditions hold.
\begin{itemize}
\item
For all $P_i,P_j \in {\cal P}$ we have $p_{i,j}\pi_i = p_{j.i}\pi_j$. 
That is, in the stationary distribution, the expected number
of transitions per unit time from state $P_i$ to state $P_j$
and from state $P_j$ to state $P_i$ are equal. For an ergodic chain,
if $\pi$ is a positive vector satisfying above condition and
the normalization condition $\sum_i \pi_i =1$, then
the chain is reversible and $\pi$ is its stationary distribution.
\item
The matrix $D^{1/2}QD^{-1/2}$ is symmetric, where $D^{1/2}$ is the
diagonal matrix $\mbox{diag}(\pi_0^{1/2} , \ldots , \pi_{N-1}^{1/2})$
and $D^{-1/2}$ is its inverse.
\end{itemize}
\begin{example}\label{ex.dictator}
\rm
We can formulate a ``dictatorial coin'' example in 
the evolutionary format of Equation~\ref{eq.Btransform}. 
Let the $B$ transformation be given by Figure~\ref{fig.Btrans}
\footnote{This $B$ gives rise to an ergodic Markov chain,
has strictly positive entries,
and it satisfies $b(u,v,w,z)=b(v,u,w,z)=b(u,v,z,w)= b(v,u,z,w)$
and hence is symmetric in the sense of
\cite{RSW92}.}
\begin{figure}
\begin{center}
 \[
 \begin{array}{cccc|c||cccc|c}
 u & v & w & z & b(u,v,w,z) &
 u & v & w & z & b(u,v,w,z) \\
 \hline
 0 & 0 & 0 & 0 & 1- 4 \epsilon &1 & 0 & 0 & 1 &   \frac{1}{2} - \epsilon\\
 0 & 0 & 1 & 0 &  \epsilon & 1 & 0 & 1 & 0 &  \frac{1}{2} - \epsilon \\
 0 & 0 & 0 & 1 &  \epsilon &1 & 0 & 1 & 1 &  \epsilon \\
 0 & 0 & 1 & 1 &  2 \epsilon & 1 & 0 & 0 & 0 &  \epsilon \\
 0 & 1 & 1 & 0 &  \frac{1}{2} - \epsilon &1 & 1 & 1 & 1 & 1 - 4 \epsilon \\
 0 & 1 & 0 & 1 &  \frac{1}{2} - \epsilon &1 & 1 & 0 & 1 &  \epsilon \\
 0 & 1 & 0 & 0 &  \epsilon &1 & 1 & 1 & 0 &  \epsilon \\
 0 & 1 & 1 & 1 &  \epsilon &1 & 1 & 0 & 0 &  2\epsilon \\
 \end{array}
 \]
\end{center}
\caption{Dictatorial $B$-transformation}\label{fig.Btrans}
\end{figure}
where $0 < \epsilon \leq \frac{1}{8}$.
The evolution of the probability densities 
by transformation of the distributions
according to Equation~\ref{eq.Btransform} 
gives
\[ p \rightarrow p' \rightarrow p'' \rightarrow \cdots \]
with $p'(0)=p'(1) = 1/2$ ($=p(0)=p(1)$) by symmetry between ``0'' and ``1.''
But if we look at what the system does in actual evolutions
then the following happens.
Consider a state space ${\cal P}$ consisting
of the two-element populations: $P_0 = \{0,1\}$,
$P_1 = \{0,0\}$ and $P_2 = \{1,1\}$. To pass from one generation to
the next one, with uniform probability and with replacement 
draw two elements from the current population.
Subsequently, execute a cross-over according to the $B$ matrix.
The resulting two individuals form the next generation.
For example,  from a population $P_0 = \{0,1\}$ we obtain
as the next generation
\begin{itemize}
\item
with probability $1/4$ a population $P_1=\{0,0\}$ with $p_1 (0)=1, p_1 (1)=0$;
\item
with probability $1/4$ 
a population $P_2=\{1,1\}$ with $p_2(0)=0, p_2(1)=1$; and
\item
with probability $1/2$ 
a population $P_3=\{0,1\} (=P_0)$ with $p_3(0)=1/2, p_3(1)=1/2$.
\end{itemize}
The associated Markov chain of this process is given by the matrix
\[
Q := \left( \begin{array}{ccc}
\frac{1}{2} & \frac{1}{4} & \frac{1}{4} \\
2 \epsilon & 1- 4\epsilon & 2\epsilon \\
2 \epsilon &  2\epsilon & 1-4 \epsilon 
\end{array}
\right)
\]
where the entry $q_{i,j}$ gives the transition probability of
going from state (population) $P_i$ to $P_j$, $0 \leq i,j \leq 2$.

Denote the probability of being
in state $P_i$ after $t$ steps by $\pi^t_i$,
and $\pi^t = (\pi^t_0, \pi^t_1, \pi^t_2)$.
Since the Markov chain is ergodic (or by simple inspection)
the pointwise limit $\lim_{t \rightarrow \infty} \pi^t \rightarrow \pi$
exists where
$\pi$ is the stationary distribution. Solving
$\pi Q = \pi$ gives 
\begin{eqnarray*}
\pi_0  =  \frac{4 \epsilon}{1+4 \epsilon} 
&\rightarrow & 0 \mbox{ for } \epsilon \rightarrow 0 \\
\pi_1  =  \frac{1}{2+8 \epsilon} 
& \rightarrow & \frac{1}{2} \mbox{ for } \epsilon \rightarrow 0 \\
\pi_2  =  \frac{1}{2+8 \epsilon} 
&\rightarrow & \frac{1}{2} \mbox{ for } \epsilon \rightarrow 0 
\end{eqnarray*}
In fact, starting from population $P_0$ after $t$ steps and
with $\epsilon$ small enough to satisfy $t \ll - \log \epsilon$ we will
be in population $P_0$ with probability
$\approx 1/2^t$, and in both population $P_1$ and
population $P_2$ with probability 
$\approx 1/2(1-1/2^t)$. Once we are in population 
$P_1$ or $P_2$ we stay in that population at the next
generation with probability $1-4 \epsilon$, for small $\epsilon > 0$
this is almost surely.
Therefore, the evolution from $P_0$ will quickly
settle in either $P_1$ or $P_2$ and henceforth remain there for a long time.

Additionally we observe that for $\epsilon = 1/8$ we have
that $Q$ is equal to its transpose $Q^T$ and
the stationary distribution $\pi = (\frac{1}{3},\frac{1}{3},\frac{1}{3})$
and therefore the chain is reversible.
\end{example}

\subsection{Finite Population: Large Sample}
Assume that we have a finite population $P$ with probability density $p(u)$
of drawing individual $u$ in the selection phase.
Assume furthermore that in the selection
phase we draw a sample of cardinality $s$.
The larger $s$ is the better we can approximate $p$ 
by the resulting frequencies.
Quantitatively this works
out as follows.

Let there be $r$ types of individuals in $\Omega$
and let $s(u,v)$ be an outcome of the random variable measuring the
number of outcomes of the pair $(u,v)$ in $s$ trials. By Chernoff's bound, 
see for example \cite{LV93},
\[ \Pr \left\{ |s(u,v)- p(u)p(v)s| > \epsilon s \right\}
< \frac{2}{e^{\alpha}} \mbox{ with } \alpha = \frac{\epsilon^2 s}{3p(u)p(v)} .\]
Let $p'()$ be the next probability distribution as
defined in Equation~\ref{eq.Btransform}, 
and let $\hat p'()$ be the frequency distribution
we obtain on the basis of the outcome $s(u,v)$ in drawing $s$ examples.
For our further considerations the dependence of $\alpha$ on $u,v$
is problematic.
It is convenient to replace $\alpha= \alpha (u,v)$ by an 
$\alpha'$ independent of $u,v$ and
$\alpha' \leq \epsilon^2 s /3 \leq \alpha (u,v)$.
Then,
\[ \Pr \left\{ |s(u,v)- p(u)p(v)s| > \epsilon s \right\}
< \frac{2}{e^{\alpha '}} \]
for every $u,v \in \Omega$.
This gives the probability that we exceed the value $\epsilon s$
for one pair $(u,v)$. The probability that we exceed the value $\epsilon s$
for {\em some} pair $(u,v)$ is upper bounded by $ 2r^2/e^{\alpha '}$.
Hence the probability that we do {\em not} exceed the value $\epsilon s$
for any pair $(u,v)$ is at least $1- (2r^2/e^{\alpha '})$.
We can now conclude that
for every $z\in \Omega$, the {\em absolute error}
of the estimate $\hat p' (z)$ is
\begin{eqnarray*}
|p'(z) -  \hat p' (z)| & \leq &  \sum_{u,v} \left(
\left| \frac{s(u,v)}{s} - p(u)p(v) \right| \sum_w b(u,v,w,z) \right) \\
& \leq & \epsilon  \sum_{u,v,w} b(u,v,w,z) = \epsilon r ,
\end{eqnarray*}
with probability at least $1-2r^2/e^{\alpha'}$.
For example, choose $\epsilon = 1/s^{1/4}$ and 
sample size 
$s$ with $s^{1/4} > 3 p(u)p(v)$ to ensure that 
$\epsilon^2 s /3 \geq \alpha' >  s^{1/4}$, for all $u,v \in \Omega$.
(The upper bound of $\alpha'$ was required by the relation
between $\alpha'$ and $\alpha$.)

Then, for all types $z \in \Omega$, for $s^{1/8} \geq r$ 
\[ \Pr \left\{|p'(z)- \hat p'(z)| <  
\frac{1}{s^{1/8}}\right\} > 1- \frac{2r^2}{e^{s^{1/4}}} 
(\geq  1- \frac{2s^{1/4}}{e^{s^{1/4}}}). \] 
That is, for growing  $s$ the probability that the 
estimator $\hat p'(z)$ differs from the real $p'(z)$
by more than $1/s^{1/8}$ decreases as $e^{-s^{1/4}}$.
We call such a sample {\em large} because it is polynomial in
the number of types $r$ which typically means that the sample is exponential
in the problem parameter $l$ (as when $\Omega = \{0,1\}^l$).
\footnote{Cubic results appearing in \cite{Ho75,Go89}
are cubic in the population size $n$ and refer to different issues.}
It is possible to estimate a boundary envelope on the
evolution trajectories around the single evolution trajectory
determined by Equation~\ref{eq.Btransform}
as a function of the sample size.
This may be the subject of a future paper.
\footnote{In a more
restricted setting of a quadratic cross-over system
with $\Omega = \{0,1\}^l$ reference \cite{RRS95}
shows that the probability distribution of an infinite
quadratic cross-over system (without fitness selection) stays
 for the duration of an evolution of $t$ generations in an
appropriate sense  close to that
of a population  of size $O(n^2 t) $ initially drawn randomly from 
the infinite population.}

\subsection{Finite Population: Small Sample}
Consider a population $P$ with types out of $\Omega$ with associated 
probability density $p( \cdot )$ of types
over $\Omega$ and local transition matrix
$B=(b(u,v,w,z))$. We generate the next population by 
drawing a {\em small} sample
consisting of {\em one} pair $u,v$ of individuals from $P$ 
(without replacement)
according to probability density $p$ and 
replace $u,v$ in $P$ by a pair $w,z$ with probability $b(u,v,w,z)$ 
to obtain population $P'$ with associated probability density $p'(\cdot )$.

In a concrete computational run of a genetic algorithm 
this means that given a population $P$ with
associated density distribution $p (\cdot )$ we obtain
with probability
$p(u)p(v)b(u,v,w,z)$ a distribution $\hat p' (\cdot)$
being the associated probability density of the population $P'$
resulting from eliminating a pair of individuals $u,v$ from $P$
and adding a pair of individuals $w,z$ as in Example~\ref{ex.small}.
Start with a population of size $n$ containing $\Omega (n)$
different types and let 
each types have positive probability to
be selected to produce the next generation. Then
there are $\Omega (n^4)$ different distributions that can be obtained
from $p(\cdot)$ this way 
(by Equation~\ref{eq.Btransform}). 

Repeating this procedure, we potentially obtain in $t$ steps up
to $n^{4t}$ distributions. For example, if $B=(b(u,v,w,z))$
is strictly positive and the associated Markov chain
is ergodic then this means that
in 
\[t_0 \approx \frac{\log N}{4 \log n} \]
generations we can possibly realize the total range of all $N$ 
different populations and therefore all 
distributions $\hat p' (\cdot)$.
Every population $P \in {\cal P}$ is obtained with some
probability in $t_0$ generations. The single
deterministic evolution of $t_0$
generations of {\em probability
distributions} 
\[p = p^0 \rightarrow p^1 \rightarrow \cdots
\rightarrow p^{t_0} \] 
according to 
Equation~\ref{eq.Btransform} gives the {\em expectation} $p^{t_0}(z)$
of an individual of type $z$ in the $t_0$th generation,
but it does not 
say anything about the actual probability density $\hat p^{t_0}(h)$
in the $t_0$th generation of an actual evolution.

\section{Towards a Discipline of Evolutionary Programming}
\label{sect.rapmix}
The upshot of the considerations so far 
is that with limited size populations and population updates
the variation in evolutions 
is very great. In practice we always deal with very
limited size populations such as, say, 500 individuals.
The question arises how to overcome the problem that
an individual evolution can become trapped in an undesirable
niche---as
in Example~\ref{ex.dictator}---for example a niche consisting
of populations with non-optimal individuals.
The answer is that we need to randomize over the
evolutions. Inspecting the populations in such a random
sample of evolutions we want to find, almost surely,
an individual of best fitness.
The latter is easy if the set of inspected evolutions
is so large that it covers almost all populations.
But this is infeasible in general. Let us look at two
easy tricks that point the way we have to go.

\begin{example}[Using the Law of Large Numbers]
\rm
Consider an ergodic Markov chain associated with an evolutionary process.
Using the law of large numbers, $c_t(P)/t \rightarrow \pi (P)$
as $t \rightarrow \infty$ almost surely, where $c_t (P)$ is the number
of occurrences of population $P$ in the first $t$ generations,
and $\pi(P)$ is the stationary probability of $P$. Therefore,
in the same manner, it is easy to show 
$\sum_{P \in {\cal P}^* }  c_t (P)/t \rightarrow \pi({\cal P}^*)$ almost
surely where
${\cal P}^*$ is the set of populations that include an individual $i^*$
with the best fitness value, and $\pi ({\cal P}^*)$ is the stationary
probability that a population includes $i^*$.

But this approach doesn't give a speed of convergence guarantee. What we 
actually want is an approach that the expected time
for an element of ${\cal P}^* $ to show up is
polynomial. One way to formulate a sufficient
condition for this is that we guaranty that for all parameters
$\epsilon, \delta > 0$ the probability
\begin{equation}\label{eq.appr}
 \Pr \left\{ \left| \sum_{P \in {\cal P}*} \frac{c_t(P)}{t} - 
\pi ({\cal P}^* ) \right| > \epsilon \right\} 
 <  \delta  , 
\end{equation}
 with $t$ polynomial in the problem parameter $l$ (like the length
of the individuals), $1/\epsilon$
and $1/\delta$.
Roughly speaking, this is achieved by ``rapidly mixing'' processes below.
\end{example}

\begin{example}[Probability Boosting]
\rm
As M.O. Rabin and others have observed, the power of randomization
over deterministic algorithms is that it can solve
problems with probability almost 1 by repeated
independent runs of the algorithm, 
provided each single run has probability of error  
appropriately less than 1 (see the text \cite{MoRa95}). 
A direct application of probability boosting
to evolutionary computing, \cite{Ru97a}, is as follows.
Let $T$ be a random variable
defined as the first-case hit
for an optimal solution by a randomized algorithm like a GA.
Let the expectation satisfy ${\bf E}(T) \leq \hat{T}$ where
the upper bound $\hat{T}$
is a polynomial in the problem dimension $l$ (like the length of
the individuals).
Now if the randomized algorithm is stopped after $t \geq 2 \hat{T}$
steps then the best solution found so far may not be the globally optimal
solution. The probability that it is not is $P(T > t)$ which
by Markov's inequality satisfies 
\[P(T>t) \leq 
\frac{{\bf E}(T)}{t} \leq \frac{1}{2}. \]
After $k$ independent runs (with independently random initial
conditions) the probability that the global solution is
found at least once is greater or equal to $1-1/2^k$.

For this observation to be useful we must
show that in the case of interest the expected 
running time up to first-case hitting time of an optimal solution
(or approximately optimal solution) is polynomial in the problem dimension.
In fact, it suffices if this is the case with respect to only a subset
of the computations of appropriate positive probability, like
in Equation~\ref{eq.appr}.
\end{example}

\subsection{Rapidly Mixing Markov Chains}\label{ap}
We follow the exposition in \cite{Si92}. Given an ergodic Markov chain,
consider the problem of sampling elements from the state
space, assumed very large, according to the stationary
distribution $\pi$. The desired distribution
can be realized by picking an arbitrary initial state
and simulating the transitions of the Markov chain according to 
probabilities $p_{i,j}$, which we assume can be computed locally
as required. As the number $t$ of simulated steps increases,
the distribution of the random variable $X_t$ will approach $\pi$.
The rate of approach to stationarity can be expressed in the 
following time-dependent measure of deviation from the limit.
For every non-empty subset $U \subseteq {\cal P}$, the {\em relative
pointwise distance} (r.p.d.) over $U$ after $t$ steps is given by
\[ \Delta_U(t) = \max_{i,j \in U} \frac{|p_{i,j}^t- \pi_j|}{\pi_j} . \]
This way, $\Delta_U(t)$ is the largest relative distance between
$\pi^t$ and $\pi$ at a state $P_j \in U$, maximized over all
possible states in $U$. The parameter $U$ allows us to specify
relevant portions of the state space. In case $U={\cal P}$ we will
omit the subscript and write $\Delta$ instead of $\Delta_U$.

The stationary distribution $\pi$ of an ergodic chain is
the left eigenvector of $Q$ with associated eigenvalue $\lambda_0 = 1$.
Let $\lambda_1 , \ldots , \lambda_{N-1} $ with $\lambda_i \in {\cal C}$
(the complex numbers) be the remaining eigenvalues (not necessarily distinct)
of $Q$. By the standard Perron-Frobenius theory for non-negative
matrices these satisfy $|\lambda_i| < 1$ for $1 \leq i \leq N-1$.
The transient behavior of the chain, and hence its rate of
convergence, is governed by the magnitude of the eigenvalues 
$\lambda_i$. In the reversible case, the second characterization
above implies that the eigenvalues of $Q$ are those of the
symmetric matrix $D^{1/2}QD^{-1/2}$ and so are all real.
This leads to the following clean formulation of above dependence:
\begin{lemma}\label{lemma.Si1}
Let $Q$ be the transition matrix of an ergodic reversible
Markov chain, $\pi$ is stationary distribution, and
$\lambda_0 =1 , \ldots , \lambda_{N-1}$ its (necessarily real) eigenvalues.
Then, for every nonempty subset $U \subseteq {\cal P}$ and all $t \in {\cal N}$
the relative pointswise distance over $U$ satisfies
\[ \Delta_U (t) \leq \frac{\lambda_{\max}^t}{\min_{P_i \in U} \pi_i} , \]
where $\lambda_{\max}$ is the largest value 
in $|\lambda_1| , \ldots, |\lambda_{N-1}|$.
\end{lemma}

\begin{lemma}\label{lem.rpd}
With the notation of Lemma~\ref{lemma.Si1} 
the relative pointswise distance over ${\cal P}$ satisfies
\[ \Delta (t) \geq \lambda_{\max}^t \]
for every even $t \in {\cal N}$. Moreover, if all
eigenvalues of $Q$ are non-negative, then
the bound holds for all $t \in {\cal N}$. 
\end{lemma}

Therefore, provided $\pi$ is not extremely small
in some state of interest, the convergence of the reversible chain
will be rapid iff $\lambda_{\max}$ is suitably bounded
away from 1. Such a chain is called {\em rapid mixing}.

 If we order the eigenvalues 
$1=\lambda_0 > \lambda_1 \geq \cdots \geq \lambda_{N-1} > -1$
then $\lambda_{\max} = \max\{\lambda_1, |\lambda_{N-1}|\}$
and the value of $\lambda_{N-1}$ is significant only if 
some eigenvalues are negative. The oscillatory behavior
associated with negative eigenvalues cannot occur
if each state is equipped with sufficiently large self-loop
probability. It is enough to have $\min_j q_{j,j} \geq 1/2$.
To see this, let $I_N$ denote the $N \times N$ identity
matrix and consider the non-negative matrix
$2Q-I_N$, whose eigenvalues
are $\mu_i = 2\lambda_i -1$. By Perron-Frobenius,
$\mu_i \geq -1$ for all $i \in {\cal P}$ which implies
that $\lambda_{N-1} \geq 0$.

What do we do when we have negative eigenvalues?
To be able to apply Lemma~\ref{lem.rpd}
without oscillations
we require all eigenvalues to be positive. It
turns out that we there is a simple modification of 
the chain with negative eigenvalues that turns it
into a chain with only positive eigenvalues without
slowing down the convergence to stationarity too much. 
We simply increase the self-loop probability of every state by $\frac{1}{2}$
after halving it first:
\begin{lemma}
With the notation of Lemma~\ref{lemma.Si1},
let the eigenvalues of $Q$ be ordered 
$1=\lambda_0 > \lambda_1 \geq \cdots \geq \lambda_{N-1} > -1$.
Then the modified chain with transition matrix
$Q'= \frac{1}{2}(I_N+Q)$, with $I_N$ as above, is also ergodic
and reversible with the same stationary distribution,
and its eigenvalues $\lambda'_i$ similarly ordered
satisfy $\lambda'_{N-1} > 0$
and $\lambda'_{\max} = \lambda'_1 =  \frac{1}{2}(1+\lambda_1)$.
\end{lemma}
Following \cite{Si92} we define rapid mixing.
\begin{definition}\label{def.rm}
Given a family of ergodic Markov chains ${\cal M}(x)$
parametrized on strings $x$ over a given alphabet.  
For each such $x$, let $\Delta^{(x)}(t)$ denote the r.p.d.
of ${\cal M}(x)$ over its entire state space after $t$
steps, and define the function $\tau^{(x)} (\epsilon )$
from the positive reals to the natural numbers by
\[\tau^{(x)} (\epsilon ) =
\min \{t: \Delta^{(x)}(t') \leq \epsilon
\mbox{ for all } t' \geq t \} . \]
We call such a family {\em rapidly mixing} iff there exist
a polynomial bounded function $q$ such that
$\tau^{(x)} (\epsilon ) \leq q(|x|, \log \epsilon^{-1} )$
for all $x$ and $0 < \epsilon \leq 1$.
\end{definition}
In the applications to evolutionary programming,
$x$ will be a problem instance and the state space of
${\cal M}(x)$ will include solution sets $R(x)$
of some relation $R$.

The question arises whether the approach to rapidly mixing Markov
chains can be generalized from {\em reversible} chains to
{\em non-reversible} chains. This was affirmatively settled in
\cite{Mi89} and another treatment was later given in ~\cite{Fi91}.
See the short discussion in \cite{Si92}.
\begin{example}
\rm
To compute the {\em permanent} of a dense matrix
is $\#P$-complete. The permanent of an $n \times n$
 matrix $A$ with 0-1 entries $a_{i,j}$
is defined by
\[ \mbox{per}{A} := \sum_{\sigma} \Pi_{i=0}^{n-1} a_{i, \sigma(i)} ,\]
where the sum is over all permutations of the set $\{0, \ldots , n \}$.
Since the class of $\#P$-complete decision
problems includes the class of NP-complete decision problems,
computing the permanent is at least NP-hard.
 
A celebrated result of Jerrum and Sinclair \cite{JeSi89} 
shows
how to use rapidly mixing Markov chains
to obtain a randomized algorithm that approximates
the value of the permanent of a matrix $A$ within ratio
$1+\epsilon$ with probability at least $3/4$ in time polynomial
in $|A|$ and $|1/\epsilon|$ where $|\cdot|$ denotes
the length of the binary representation. 
By probability boosting, we can by $O( \log \delta )$  iterations
boost the success probability to at least $1- \delta$.
This breakthrough result has led to a ``Markov Chain Renaissance''
to employ rapidly mixing Markov chains to obtain such
``fully polynomial randomized approximation schemes ({\em fpras})'' to
hard problems in computer science \cite{DiHo95,Si92,ADSS95}.
These applications generate a uniform stationary distribution
from which the approximation is obtained by Monte Carlo sampling of the
states and determining the proportion of the successful states.
In our application to genetic computing we proceed differently:
with high probability we sample states
containing a best fit individual (or an approximately best fit
individual). 
We illustrate the idea by example in Section~\ref{sect.toy}.
\end{example}

\subsection{Optimization by Rapidly Mixing Evolutionary Algorithms}
To optimize by rapidly mixing evolutionary algorithms
we require two properties:
\begin{enumerate}
\item\label{prop.concentrate}
The stationary distribution $\pi$ of populations ${\cal P}$
of the associated
Markov chain of the evolutionary process converges to
concentrate a sufficient amount of probability on
populations containing maximally fit individuals, or
on populations containing individuals that enable us to compute the
required solutions.
That is, 
\[\sum_{P_i \in {\cal P}^*} \pi_i \geq \epsilon ,\]
where ${\cal P}^*$ is the set of populations containing at
least one solution of best fitness (or a solution that
approximates the global optimum, or solutions that enable us to compute the
required solution or approximation) 
 and $\pi_i$ is the
stationary probability of population $P_i$. To ensure
feasibility of the algorithm we customarily
 require that $1/\epsilon$ is polynomial in the problem parameter.
\item\label{prop.rapmix}
The Markov chain of the evolutionary process converges
sufficiently fast to the stationary distribution:
it is {\em rapidly mixing} as in Section~\ref{ap}.
\end{enumerate}

The rapid mixing property~(\ref{prop.rapmix})
can be satisfied by having the evolutionary system
satisfy some structural properties.
Such properties can, at least in principle (if not
in practice), always be taken
care of while implementing the evolutionary system
by choosing the selection rules, cross-over operator,
and mutation rules appropriately. These requirements
are covered in Section~\ref{sect.struct}.

The question of probability concentration, property~(\ref{prop.concentrate}),
is more subtle, and it is not yet clear how to
generally go about it, even in principle. 
In many if not most cases we are satisfied to
obtain an {\em approximately} globally optimal solution.  

\subsection{A Discipline of Evolutionary Programming}\label{sect.struct}
For a structural discipline of evolutionary programming
we need to develop a methodology that given a problem
specification guides us to construct an evolutionary system
such that the associated Markov chain satisfies the following requirements:
\begin{property}
\label{it.2nd}
the second largest eigenvalue\footnote{The second largest eigenvalue
was used earlier in genetic computing 
to advantage for another purpose in \cite{Su93}.}
$\lambda_{\max}$ is suitably bounded away far enough from 1 so that the
Markov chain is rapidly mixing (Definition~\ref{def.rm} of
Section~\ref{ap}); and
\end{property}
\begin{property}
\label{it.invpol}
the stationary distribution $\pi$ gives probability greater than $\epsilon$,
where $1/\epsilon$ is polynomial in the problem parameter,
to the set of states that contain 
individuals of best fitness.
\end{property}

For Property~\ref{it.2nd} it is required that the matrices
are (i) irreducible, and (ii) have nonnegative entries.
Since the only matrices we consider are {\em stochastic}
where the entries are transition probabilities,  
(ii) is in our case easy to satisfy up to the `suitable' condition
in Property~\ref{it.2nd}. 
Since we only deal
with ergodic matrices, and (i) is required for ergodicity,
Property~\ref{it.2nd} is always satisfied in our case. Ergodicity
is immediate if we have a positive mutation probability of
transforming $i$ into $j$ for each pair of types $i,j$.
Hence by proper choice
of the genetic system leading to suitable
transition probabilities inducing 
a rapidly mixing Markov chain 
one can satisfy Property~\ref{it.2nd} in construction
of an evolutionary system. It is perhaps less easy
to see whether
it is feasible to satisfy Property~\ref{it.invpol} in each 
particular case, or indeed without knowing the optimal individual
a priori. However, as discussed above a similar approach for approximating very hard
combinatorial optimization problems, \cite{Si92},
worked out fine.

Assume that we have defined our evolutionary system satisfying
Properties~\ref{it.2nd}, \ref{it.invpol}. The program we use is then as follows.
Repeat a polynomial number of times:
\begin{description}
\item[Step 1:]
From a start state evolve through a polynomial number of
generations;
\item[Step 2:]
From the final population vector select the fittest individual.
\end{description} 
%

{\bf Paradigm.} Running the
program longer than a polynomial number of generations
 will not significantly change
the closeness of the state distribution to the stationary
distribution in the Markov chain. We can only guarantee
that we find a state (vector) containing an optimal fit
individual with probability say inversely polynomial
in the problem parameter. However, polynomially
repeating this procedure implies Monte Carlo sampling 
which almost surely discovers the individual with
optimal fitness.

\section{A Toy Rapidly Mixing Genetic Algorithm}\label{sect.toy}
Consider a toy evolutionary problem as follows.
We consider a population of size $\sqrt{l}$ and very simple crossover only
and some mutation.
This example already 
illustrates adequately the rapid mixing phenomenon.
 The genetic algorithm $G$ is defined as follows. The set of
all program types is $\Omega = \{0,1\}^l$ with $l$ fixed, even,
and large enough for the following analysis to hold. The {\em fitness}
of a program $\omega \in \Omega$ with $\omega =\omega_1\omega_2 \ldots \omega_l$
is given by the function 
\[ f(\omega)=1 \mbox{ if } \sum_{i=1}^l \omega_i = l/2, \mbox{ and } 1/2 \mbox{ otherwise }. \]
The {\em starting population} $P^0$ at time $t_0 = 0$ contains
$\sqrt{l}$ copies of the individual $00\ldots 0$;
its cardinality (number of elements in $P^0$) is $\sqrt{l}$.
We express the {\em frequency} of
a string $\omega$ in a population $P$ by $\#_\omega (P)$.
That is, 
$\#_{00 \ldots 0} (P^0)= \sqrt{l}$
and
$\#_{\omega} (P^0)= 0$ for
$\omega \neq 00 \ldots 0$

The transition of one population to the next generation
(population) is as follows.
To avoid problems of periodicity, we add self-loop probability
of $1/2$ to each state (that is, population). Note that this also
dispenses with the problem of negative eigenvalues.
 Consequently, there is probability $1/2$
that the state changes using crossover and mutation,
and there is probability $1/2$ that it stays the same.
The probability $p(\omega)$
of selecting a string $\omega$ from a population $P$ is
\begin{equation}\label{eq.pr}
p(\omega)= \frac{\#_\omega (P) f(\omega)}{\sum_{\omega \in \Omega} \#_\omega (P) f(\omega)} .
\end{equation}
In the selection phase we select two individuals in $P$, say $\omega^i,
\omega^j$,
according to these probabilities,
and with probability 1/2 we perform a crossover and mutation on each
(and with probability 1/2 we do nothing). 
The crossover operator interchanges a single bit
of $\omega^i$ with the corresponding bit of $\omega^j$. It selects the single bit
position with uniform probability $1/l$. 
Subsequently, we mutate each offspring by flipping
a single bit with uniform probability
$ 1/l$ chosen from the positions $1$ through $l$.
(If $i=j$ then the cross-over
doesn't do anything and the two mutations may result in 0,1, or 2 bit flips
of $\omega_i$.)
We first prove that $G$ is rapid mixing by showing that if
the following system $G'$ is rapidly mixing then so is $G$.

Let $G'$ be a system where the initial state is
a binary $l$-vector. At each step uniformly
at random select a bit position of the current $l$-vector and flip that bit
with fifty-fifty probability to produce the next $l$-vector.
Then $G'$ is a Markov chain where the states are the binary $l$-vectors.

\begin{lemma}\label{lem.sincl}
The chain $G'$ is rapid mixing with r.p.d. at most $\epsilon$
within $O(l^{2} (l + \log ( 1/ \epsilon)))$ steps.
\end{lemma}
For a proof see \cite{Si92}, pp. 63--66.
This system is an almost uniform generator for $\Omega$,
using singleton populations, where it suffices
to use  an  arbitrary starting
singleton population.
In terms of GA's it is single-bit mutation. Our example
involves single-bit mutation, single-bit cross-over,
and selection. The reader is advised that this is only a cosmetic change
to make the example look more like a `realistic' GA. 
Our toy example $G$ is
essentially the example $G'$ as in Lemma~\ref{lem.sincl}.
To see this,
consider the vectors in successive generations $P^0,P^1, \ldots$
of $G$
to maintain their identity. If $P^t=\{\omega^{t,1},
\ldots, \omega^{t, \sqrt{l}} \}$ for $t>0$ and in the selection
phase we select indices $i,j$, then
$\omega^{t+1,k} = \omega^{t,k}$ for $0 \leq k \leq \sqrt{l}$
and $k \neq i,j$, or $\omega^{t+1,h}$ 
results from $\omega^{t,h}$ (the `same vector') by at
most two bit flips for $h=i,j$.

\begin{lemma}\label{lem.run}
Let $\epsilon > 0$ and $T(l)=O(l^{5/2} (l + \log ( 1/ \epsilon)))$.
For each $t \geq T(l)$, with probability at least $1-1/T(l)$ and
for each $l$-vector $\omega$,
every $l$-vector $\omega^{0,j} \in P^0$ has probability 
$(1\pm \epsilon) /2^l$ 
of being changed into $\omega^{t,j}=\omega$
in $t$ generations of $G$.
\end{lemma}
\begin{proof}
For a fraction of at least $1-1/t$ of all
runs of $t> \sqrt{l}$ steps of a population of $\sqrt{l}$ 
elements, then each element $j$ out of $1, \ldots , \sqrt{l}$
(representing the vector $\omega^{\cdot,j}$)
is selected with frequency 
of at least 
\begin{equation}\label{eq.rrrr}
\frac{t}{2\sqrt{l}} \pm O( \sqrt{ \frac{t \log t}{\sqrt{l}}} )
\end{equation}
in the selection phases of the generating process.
This is shown similar
to the statistical analysis of `block frequencies'
of high Kolmogorov complexity strings in \cite{LV93},
Theorem 2.15. 

Namely, consider $t$ throws of a $\sqrt{l}$-sided coin,
each pair of throws constituting the selection of the indexes
of the two individuals mated to produce the next generation.
There are $2^{(t \log l)/2}$ possible sequences $x$ of $t$ outcomes. Hence, the
maximal Kolmogorov complexity is given by $C(x|t,l) \leq (t \log l)/2 +O(1)$.
Moreover, since there are
only $2^{(t \log l)/2}/t$ binary descriptions of length
$< (t \log l)/2 - \log t +O(1)$,
there is a fraction of at least $1-1/t$th part of all sequences $x$
which has $C(x|t,l) \geq (t \log l)/2 - \log t +O(1)$.
Consider each such $x$ as a binary string consisting
of blocks of length $\log \sqrt{l}$, each block encoding
one of the $\sqrt{l}$ types.
Let $\#j(x)$ denote the number of occurrences of each of
the $\sqrt{l}$ blocks $j$ (elementary outcomes) 
in $x$. Then, by \cite{LV93} p. 163, 
\[|\#j(x) - t/\sqrt{l}| \leq 
\sqrt{ \frac{ \log \sqrt{l} + \log \log \sqrt{l} + \log t + O(1)}
{\sqrt{l}\log e}
3t.}
\]
Since
individuals have fitness $1/2$ or $1$,
some indexes may at various times 
have as low as half the probability of being selected
than other individuals. Repeating the same argument 
for an $2 \sqrt{l}$-sided coin and represent by the first $\sqrt{l}$
outcomes for indexes $1$ through $\sqrt{l}$ and 
the remaining outcomes represent dummy indexes (possibly the original ones)
we obtain the lower bound of Equation~\ref{eq.rrrr}. 

Following the same vector in the successive generations, consider each time
it is selected. At such
times, with fifty-fifty
probability either nothing is done or the vector incurs (i) a
bit flip in a position which was selected uniformly at random
because of the cross-over (or no bit flip if the bits in that
position of the two parents happened to be the same),
followed by (ii) a bit flip in a position selected  uniformly 
at random because of the mutation. From the viewpoint of the
individual vector and the mutation operations 
alone it simply emulates a trajectory
of the singleton $l$-vector in Lemma~\ref{lem.sincl}
of length as
given in  Equation~\ref{eq.rrrr}. The extra random bit flips
due to the cross-over only increase the length of the emulation.

Substitute $t$ in  Equation~\ref{eq.rrrr} by $T(l)$ as in the statement
of the lemma. By Lemma~\ref{lem.sincl} the lemma is proven.
\end{proof}

Let $\omega$ be an $l$-vector. For every $\epsilon > 0$
and $t \geq T(l)$, every $l$-vector 
in the initial population $P^0$ turns into $\omega$ in exactly
$t$ steps with probability at least $(1-1/t)(1 \pm \epsilon)/2^l$.
Therefore, $P^0$ generates in $t$ steps every particular 
population $P$ of $\sqrt{l}$ individuals with probability
\[ (1-\frac{1}{t}(1 \pm \epsilon))/N , \]
where $N$ is the number of $\sqrt{l}$-size populations.
Then, the r.p.d. of $G$ to the uniform stationary distribution $\pi$ 
($\pi (P) = 1/N$ for all $P \in \{0,1\}^l$ with $\#(P)= \sqrt{l}$)
 after $t > T(l)$ steps is bounded above by $\frac{1}{t}(1 + \epsilon)$.
Choosing $t > \max\{T(l), 1/\epsilon\}+1$ the r.p.d. is upper bounded
by $\epsilon$.

\begin{corollary}
It follows that $G$ is a rapidly mixing 
Markov Chain with a uniform stationary distribution.
\end{corollary}
\begin{lemma}\label{lem.concentr}
The probability of finding
a population with an optimally fit element in $t$
runs is at least $1- 2 e^{-\alpha t}$ with $\alpha = c /(16(1-c))$, for
the fixed constant $c$ given in Equation~\ref{eq.ccc}. 
\end{lemma}
\begin{proof}
There are ${l \choose l/2} \approx 2^l/ \sqrt{\pi l/2}$  
strings with fitness 1.
Hence a fraction of at most
 $$(1 - 1/ \sqrt{\pi l/2})^{\sqrt{l }}
< e^{- \sqrt{2/\pi}}$$ 
populations
of size $\sqrt{l}$ contain no such strings. This means that a constant
fraction of at least 
\begin{equation}\label{eq.ccc}
c= 1-e^{- \sqrt{2/ \pi}},
\end{equation}
 of the populations
of size $\sqrt{l}$ contain at least one string of fitness 1.

Consider each run of $T(l)$ generations an experiment with
a {\em success} outcome if the final population contains 
an individual with fitness 1. Let the number of successes
in $t$ trials be $s(t)$. Then, with $\beta$ defined as
\[ \beta = \Pr\{  |s(t) - c t|  > \delta t \} \]
we have
\[ \beta < 2 e^{-\delta^2 t /(3c)} , \]
by Chernoff's bound. For $\delta = c/2$ we know that the
number of successes $s(t)>0$ with probability at least $1 - \beta$.
\end{proof}

\begin{theorem}[(Rapidly Mixing GA Algorithm)]
Let $\epsilon$ and $T(l)$ be as in Lem\-ma~\ref{lem.run}
and let $\alpha$ be as in Lemma~\ref{lem.concentr}.
Repeat $t$ times: run $G$ for 
$T(l)$ generations.
This procedure uses $O(T(l) \cdot t)$ elementary
steps consisting of the generation from one population to 
the next population. (With $t=l$ this
is a low degree polynomial in $l$ and $\epsilon$). 
The probability
of finding an optimal element exceeds $$1-2e^{-\alpha t},$$
where $ \alpha > 0$, that
is, with probability of failure which vanishes exponentially
fast with rising $t$.
\end{theorem}
\begin{proof}
By Lemmas~\ref{lem.run}, \ref{lem.concentr}.
\end{proof}

\section{Non-uniform Stationary Distributions}
In the above
example the stationary distribution is uniform
and success of the method depends on the abundance
of populations containing an optimal individual.
However, we want the stationary distribution of populations to
heavily concentrate probability on populations containing
optimal or near-optimal individuals even if those populations are scarce. 
For example, if our fitness function is $f: \Omega \rightarrow {\cal N}$
and we extend $f$ to populations $P$ with 
$f(P) = \max_{\omega \in P} \{f(\omega)\}$
then we want to generate a random element from a distribution
concentrated on the set of optimum solutions. This is similar to
generating a random $P$ from a distribution $\pi$ where
$\pi (P) = \Theta (2^{f(P)/\alpha})$ with $\alpha$ a small positive 
number. Then, with large probability a random $P$ will maximize $f$.
A general method to modify a random walk so that it converges to an arbitrary
prescribed probability distribution is the {\em Metropolis filter},
\cite{MRRTT53}. Let's explain a simple example of this. Suppose
$\Omega = \{0,1\}^l$, we are dealing with singleton populations,
and our fitness function is $f$. We describe a random walk
on $\Omega$ by single bit-flips and ``filtered'' by the function $f$. 
The next population
is generated from the current population $\{\omega\}$  as follows. First
select a random bit position in $\omega$. Let $\omega'$ be the string
resulting from flipping that bit of $\omega$
If $f(\omega ') > f(\omega)$ then the next generation is $\{ \omega '\}$;
otherwise the next population is $\{\omega '\}$ with probability
$f(\omega')/f(\omega )$ and the next population is $\omega$ with
probability $1-f(\omega')/f(\omega )$. 
Clearly this modified random
walk is a Markov chain (and it is also time-reversible).
The stationary distribution $\pi^f$ is
\[ \pi^f (\omega) = \frac{f(\omega)}{ \sum_{\omega' \in \Omega} f(\omega' )}.
\]
For example, with $f( \omega ) = 2^{i^2}$ where $i$ is the number
of 1's in $\omega$ the optimal individual is $11 \ldots 1$
which is sampled from the stationary distribution with high
probability. Unfortunately, it is not in general known how to 
estimate the mixing time of a Metropolis-filtered random walk.
On the positive side,
in \cite{AK91} they compute a volume in $n$-space using this method
and they show that the filtered walk mixes essentially as fast as the
corresponding unfiltered walk. 
A similar approach to combinatorial optimization using the Markov chain Monte
Carlo method in the sense of a Metropolis process-type Markov chain
having a stationary distribution that concentrates high probability
on the optimal (or approximately optimal) solutions is surveyed
in \cite{JeSi95}. They give a polynomial time Metropolis process
to find an approximate maximum matching in arbitrary graphs 
with high probability. More precisely, if $G$ is an 
arbitrary graph on $n$ vertices then the algorithm finds a matching in $G$
of size at least $\lfloor (1- \epsilon ) k_0 \rfloor$ where $k_0$
is the size of the maximal matching and $\epsilon$ is an accuracy
parameter which is assumed to be constant---the running time is actually
exponential in $1/\epsilon$. However, these successes are scarce.
 For the current status and references on
Metropolis algorithms see \cite{DiSC95}.

\section{Conclusion and Further Research}
We have suggested a theoretical possibility of constructing genetic
processes that provably optimize an objective function 
with high probability in polynomial time. We have given
a simple example that, however, succeeds because of the abundance
of optimal solutions. 
Altogether it seems difficult at this time to even construct an
example of a genetic process that is both rapidly mixing
and also has a nonuniform stationary distribution
that heavily concentrates probability on populations containing
optimal individuals in case such populations are scarce.
An example of this would give evidence of the power of the
proposed method.


\section*{Acknowledgment}
I thank the anonymous referees and John Tromp for their helpful comments.

\end{document}